\documentclass[10pt,twocolumn,letterpaper]{article}

\usepackage{wacv}
\usepackage{times}
\usepackage{epsfig}
\usepackage{graphicx}
\usepackage{amsmath}
\usepackage{amssymb}
\usepackage{booktabs}
\usepackage{enumitem}
\usepackage{ulem}
\usepackage{makecell}
\usepackage{mathtools}
\usepackage{multirow}

\usepackage[accsupp]{axessibility}

\def\wacvPaperID{704} 

\wacvapplicationstrack 

\wacvfinalcopy 

\ifwacvfinal
\usepackage[breaklinks=true,bookmarks=false]{hyperref}
\else
\usepackage[pagebackref=true,breaklinks=true,colorlinks,bookmarks=false]{hyperref}
\fi

\begin{document}

\title{DigiFace-1M: 1 Million Digital Face Images for Face Recognition}

\author{Gwangbin Bae\\
University of Cambridge\\
{\tt\small gb585@cam.ac.uk}
\and
Martin de La Gorce\\
Microsoft\\
{\tt\small madelago@microsoft.com}
\and
Tadas Baltru\v{s}aitis\\
Microsoft\\
{\tt\small tabaltru@microsoft.com}
\and
Charlie Hewitt\\
Microsoft\\
{\tt\small chewitt@microsoft.com}
\and
Dong Chen\\
Microsoft\\
{\tt\small doch@microsoft.com}
\and
Julien Valentin\\
Microsoft\\
{\tt\small juvalen@microsoft.com}
\and
Roberto Cipolla\\
University of Cambridge\\
{\tt\small rc10001@cam.ac.uk}
\and
Jingjing Shen\\
Microsoft\\
{\tt\small jinshen@microsoft.com}
}

\maketitle
\thispagestyle{empty}

\begin{abstract}
State-of-the-art face recognition models show impressive accuracy, achieving over 99.8\% on Labeled Faces in the Wild (LFW) dataset. Such models are trained on large-scale datasets that contain millions of real human face images collected from the internet. Web-crawled face images are severely biased (in terms of race, lighting, make-up, etc) and often contain label noise. More importantly, the face images are collected without explicit consent, raising ethical concerns. To avoid such problems, we introduce a large-scale synthetic dataset for face recognition, obtained by rendering digital faces using a computer graphics pipeline\footnote{DigiFace-1M dataset can be downloaded from \url{https://github.com/microsoft/DigiFace1M}}. We first demonstrate that aggressive data augmentation can significantly reduce the synthetic-to-real domain gap. Having full control over the rendering pipeline, we also study how each attribute (e.g., variation in facial pose, accessories and textures) affects the accuracy. Compared to SynFace, a recent method trained on GAN-generated synthetic faces, we reduce the error rate on LFW by 52.5\% (accuracy from 91.93\% to 96.17\%). By fine-tuning the network on a smaller number of real face images that could reasonably be obtained with consent, we achieve accuracy that is comparable to the methods trained on millions of real face images.
\end{abstract}

\section{Introduction}
\label{sec:introduction}

\begin{figure*}[t]
\begin{center}
\includegraphics[width=\linewidth]{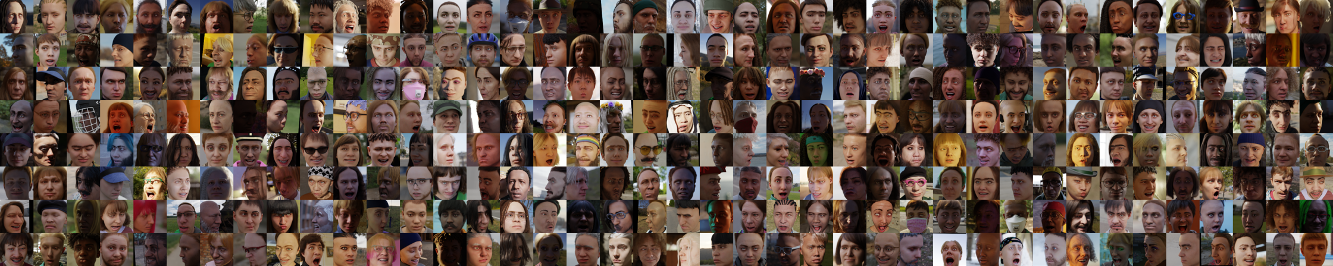}
\end{center}
\caption{Examples of synthetic face images in our dataset. Our dataset captures a wide variety of facial geometry, pose, textures, expressions, accessories and environments.}
\label{fig:intro}
\end{figure*}

Learning-based face recognition models \cite{2015_FR_facenet,2017_FR_SphereFace,2018_FR_CosFace,2018_FR_SV-AM-Softmax,2019_FR_ArcFace,2020_FR_curricularface,2021_FR_magface,2022_FR_adaface} use Deep Neural Networks (DNNs) to encode the given face image into an embedding vector of fixed dimension (e.g., 512). These embeddings can then be used for various tasks, such as face identification (who is this person) and verification (are they the same person). To learn diverse, discriminative embeddings, the training dataset should contain a large number of unique identities. To learn robust embeddings, i.e., which are not sensitive to the changes in pose, expression, accessories, camera and lighting, the dataset should also contain a sufficient number of images per identity with these variations.

Publicly available face recognition datasets satisfy both. MS1MV2~\cite{2019_FR_ArcFace} contains 5.8M images of 85K identities (approx.~68 images per ID). 
Recently released WebFace260M~\cite{2021_FRD_webface260m} contains 260M images of 4M identities (approx.~65 images per ID). 
While such datasets have driven recent advances in face recognition models, there are several problems associated with them.

\noindent{\textbf{(1) Ethical issues.}} 
Large-scale face recognition datasets are often criticized for ethical issues including privacy violation and the lack of informed consent. For example, datasets like~\cite{2014_FRD_casiawebface,2016_FRD_ms1m,2019_FR_ArcFace,2021_FRD_webface260m} are obtained by crawling web images of celebrities \textit{without consent}. To increase the number of identities, some datasets exploited the term ``celebrities'' to include anyone with online presence. Datasets like \cite{2016_FRD_megafacev1,2017_FRD_megafacev2} collected face images of the \textit{general public} (including children) from Flickr~\cite{flickr}. Projects like MegaPixels~\cite{megapixels} are exposing the ethical problems of such web-crawled face recognition datasets. Following severe criticism, public access to several datasets has been removed \cite{nature_article}.

\noindent{\textbf{(2) Label noise.}} Web images collected by searching the names of celebrities often contain label errors. For example, the Labeled Faces in the Wild (LFW) dataset~\cite{2008_FRD_LFW} contains several known errors including: (1) mislabeled images; (2) distinct persons with the same name labeled as the same person; and (3) the same person that goes by different names labeled as different persons. 

\noindent{\textbf{(3) Data bias.}} 
Face recognition models are generally trained and tested on celebrity faces, many of which are taken with strong lighting and make-up. Celebrity faces also have imbalanced racial distribution (e.g., 84.5\% of the faces in CASIA-WebFace~\cite{2014_FRD_casiawebface} are Caucasian faces~\cite{2019_FRD_RFW}), leading to poor recognition accuracy for the under-represented racial groups~\cite{2019_FRD_RFW}.

In order to circumvent all these issues that affect the existing real face datasets, we introduce a new large-scale face recognition dataset consisting only of photo-realistic \textit{digital face images} rendered using a computer graphics pipeline and make this dataset available to the community. Specifically, we build upon the face generation pipeline introduced by Wood et al.~\cite{2021_FakeItMakeIt}, tailoring the amount of variability for each attribute (e.g., pose and accessories) for our recognition task, and generate 1.22M images with 110K unique identities. Each identity is generated by randomizing the facial geometry and texture as well as the hair style. The generated face is then rendered with different poses, expressions, hair color, hair thickness and density, accessories (including clothes, make-ups, glasses, and head/face wear), cameras and environments, to encourage the network to learn a robust embedding. \autoref{fig:intro} shows examples of synthetic face images in this new dataset. We generated 1.22M images, but in practice the number of identities and images you can generate with synthetics pipeline is only limited by the cost of generating and storing these images.

Digital synthetic faces can solve the aforementioned problems associated with the real face datasets. Firstly, the generated faces are free of label noise. Secondly, the bias in lighting, make-up and skin color can be reduced as we have full control over those attributes. Most importantly, the face generation pipeline does not rely on any privacy-sensitive data obtained without consent.

This is a critical difference from the GAN-generated synthetic faces; face GANs rely (either directly or indirectly) on large-scale real face datasets to train some components of their pipeline, leaving unresolved ethical problems. For example, a recent method called SynFace~\cite{2021_FR_synface} was trained on synthetic faces generated using DiscoFaceGAN~\cite{2020_discofacegan}. While the generated face images are free of label noise, millions of real face images were used for training DiscoFaceGAN. The GANs may also inherit any bias that exists in the real face images used to train them. For our dataset, only 511 face scans, \textit{obtained with consent}, were used to build a parametric model of face geometry and texture library~\cite{2021_FakeItMakeIt}. 
From this limited source data, we can generate infinite number of identities, making our approach easily scalable.

Our contributions can be summarized as below:

\begin{itemize}[leftmargin=*]
    \item We release a new large-scale \textit{synthetic} dataset for face recognition that is free from privacy violations and lack of consent. To the best of our knowledge, our dataset, containing 1.22M images of 110K identities, is the largest public synthetic dataset for face recognition.
    \item Compared to SynFace~\cite{2021_FR_synface}, which is trained on GAN-generated faces, we reduce the error rate on LFW by 52.5\% (accuracy from 91.93\% to 96.17\%). For five popular benchmarks~\cite{2008_FRD_LFW,2016_FRD_CFPFP,2018_FRD_CPLFW,2017_FRD_AGEDB,2017_FRD_CALFW}, the average error rate is reduced by 46.0\% (accuracy from 74.75\% to 86.37\%).
    \item We demonstrate how the proposed synthetic dataset can be used in conjunction with a small number of real face images to substantially improve the accuracy. This simulates a scenario where a small number of curated (i.e., no label noise and reduced bias) real face images are collected with consent. By fine-tuning our network with only 120K real face images (i.e., 2\% of the commonly-used MS1MV2 dataset~\cite{2019_FR_ArcFace}), we achieve 99.33\% accuracy on LFW and 93.61\% on average across the five benchmarks, which is comparable to the methods trained on millions of real face images.
    \item Having full control over the rendering pipeline, we perform extensive experiments to study how each attribute (e.g., variation in facial pose, accessories and textures) affects the face recognition accuracy. 
\end{itemize}

\section{Related Work}

\noindent
\textbf{Face recognition datasets with real face images.} Major tech companies can utilize private data to train their face recognition models. Google used 100M-200M images of 8M identities to train FaceNet~\cite{2015_FR_facenet}, and Facebook used 500M images of 10M identities~\cite{2015_FR_facebook}. It is challenging to construct datasets of comparable size using face images that are publicly available. Public datasets generally rely on celebrity images~\cite{2008_FRD_LFW,2014_FRD_casiawebface,2016_FRD_ms1m,2021_FRD_webface260m} or web images that are posted with Creative Commons license~\cite{2016_FRD_megafacev1,2017_FRD_megafacev2}. As discussed in \autoref{sec:introduction}, such datasets have ethical issues and suffer from label noise and data bias.

\noindent
\textbf{Synthetic faces generated using deep generative models.}
Deep generative models such as GANs~\cite{2014_gan} can produce photo-realistic images and have been used to generate synthetic data to train face recognition \cite{2021_Trigueros, 2021_FR_synface}. While traditional generators (e.g., \cite{2017_progressive_gan}) generate a face image from a single latent vector that changes both the identity and its appearance, DiscoFaceGAN~\cite{2020_discofacegan} learned \textit{disentangled} latent representations for identity, pose, expression and illumination. SynFace~\cite{2021_FR_synface} used DiscoFaceGAN to generate a synthetic dataset for face recognition, consisting of 10K identities and 500K images. SynFace achieved 91.93\% accuracy on LFW dataset~\cite{2008_FRD_LFW}, and by mixing the synthetic dataset with 2K real identities (20 images each), the accuracy was pushed up to 97.23\%. However, their performance is poor for large-pose-variation datasets (e.g., 75.03\% on CFP-FP~\cite{2016_FRD_CFPFP} and 70.43\% on CPLFW~\cite{2018_FRD_CPLFW}). This is mainly because it is challenging to train a 2D GAN to produce images that preserve 3D geometric consistency~\cite{2022_gram}.

\noindent
\textbf{Synthetic faces generated using 3D parametric models.} Classical 3D parametric face models such as morphable models~\cite{Blanz1999AMM} explicitly model the identity independently from other parameters which makes them well suited for generating face recognition datasets. However, previous results obtained with this kind of synthetic images have shown limited performance~\cite{2019_Kortylewski,2018_undo_bias} unless combined with a large number of real images. This can be due to the lack of realism and variability in the models that have been used to generate the faces.
Wood et al.~\cite{2021_FakeItMakeIt} introduced a pipeline for generating and rendering diverse and photo-realistic 3D face models. A generative face model, learned from the 3D scans of 511 individuals, is used to generate a random 3D face. The face is then combined with artist-created assets (e.g., texture, hair, accessories) and is rendered under a random environment (simulated with HDRIs - high dynamic range images). The rendered synthetic face images (and the corresponding auto-generated ground truth annotations) were used to learn various face analysis tasks such as face parsing~\cite{2021_FakeItMakeIt}, landmark localization~\cite{2021_FakeItMakeIt,2022_dense_landmarks_are_all_u_need} and face reconstruction~\cite{2022_dense_landmarks_are_all_u_need}, demonstrating state-of-the-art performance. In this paper, we aim to demonstrate that such photo-realistic rendered synthetic faces can be used to tackle face recognition.

\section{Digital Faces for Face Recognition}

This section explains how the proposed dataset is generated.
We first explain how digital faces are controlled, rendered and aligned to create the dataset (\autoref{sec:method1}). 
After providing the dataset statistics (\autoref{sec:method3}), we introduce the data augmentation details which help in minimizing the synthetic-to-real domain-gap (\autoref{sec:method4}).

\subsection{Face Rendering}
\label{sec:method1}

We build upon the face generation and rendering pipeline introduced by Wood et al.~\cite{2021_FakeItMakeIt}. 
In this section, we explain the modifications we made to the original pipeline to create a large-scale dataset for face recognition.

\begin{figure}[t]
\begin{center}
\includegraphics[width=\linewidth]{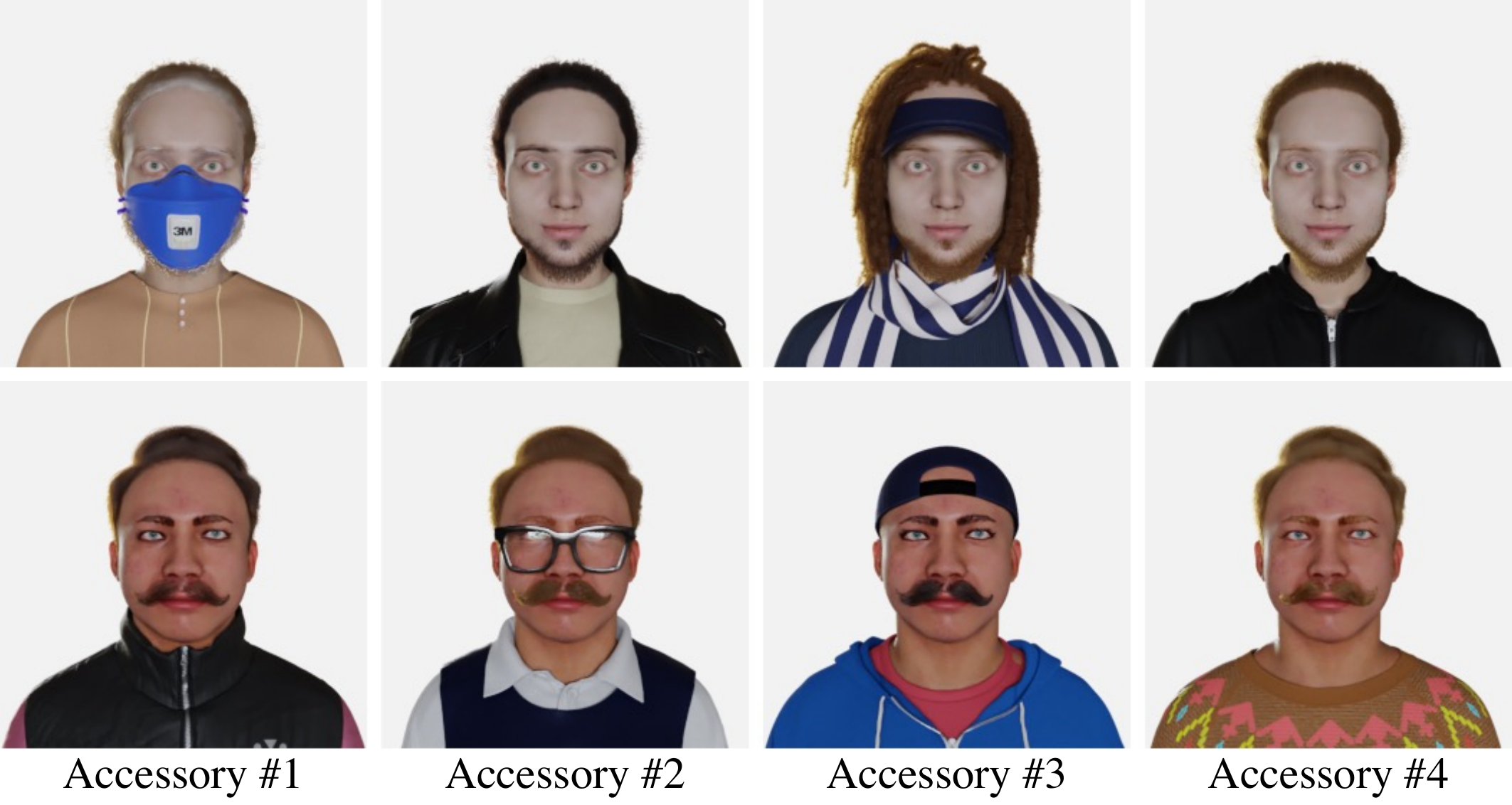}
\end{center}
\caption{Each row shows the same identity rendered with different accessory setups. Accessories include clothes, glasses, make-up (e.g., eyeshadow and eyeliner), face-wear and head-wear. The color, density and thickness of facial and head hair are also randomized. The hair style is modified only when the sampled accessory conflicts with the original hair style.}
\label{fig:id_and_acc}
\end{figure}

\begin{figure}[t]
\begin{center}
\includegraphics[width=\linewidth]{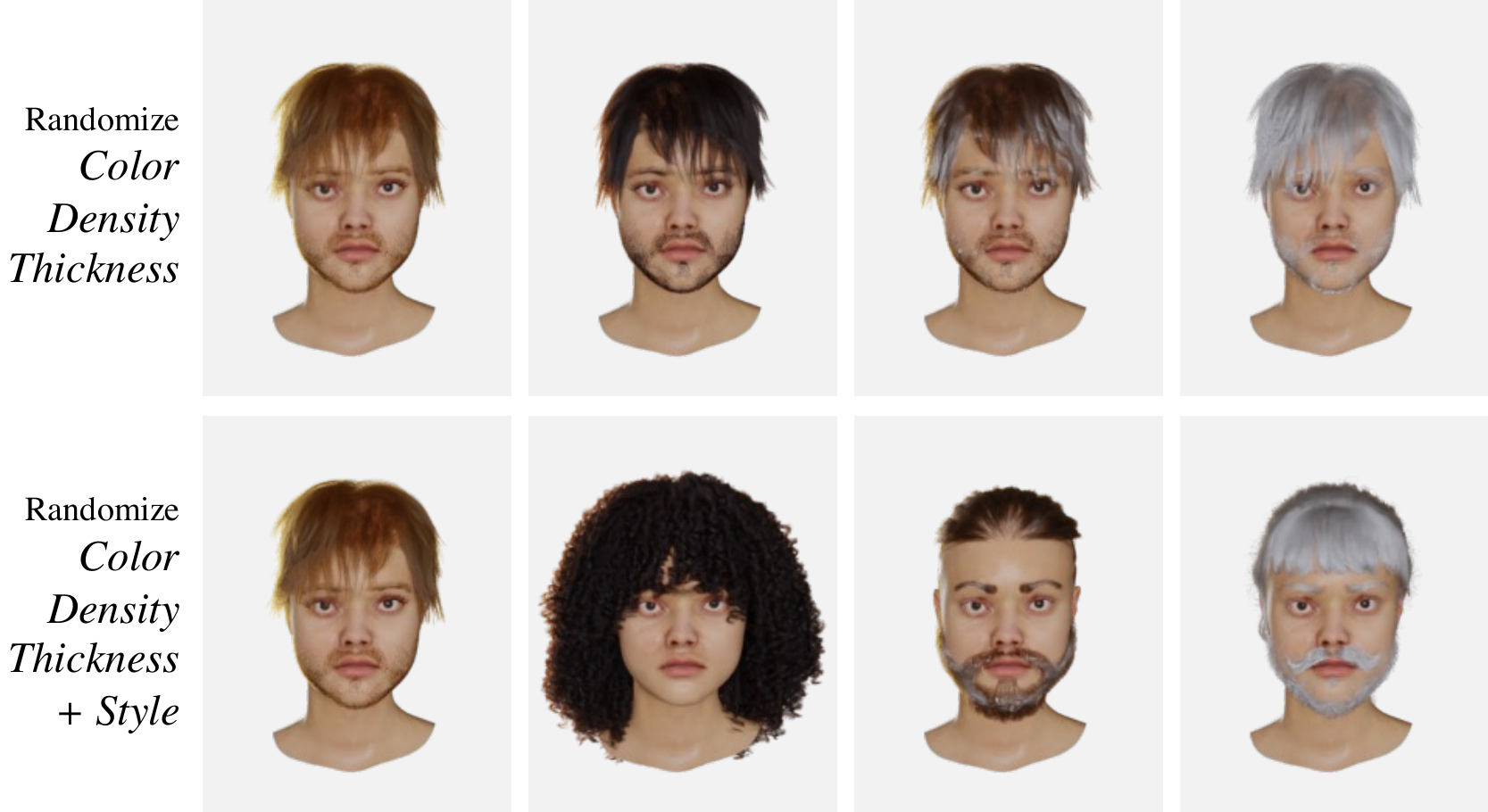}
\end{center}
\caption{Randomizing the hair style makes the problem unnecessarily difficult (see the bottom row), as most people maintain similar hair styles. Therefore, we only randomize the color, density and thickness of the hair as shown in the top row (the hair is also randomly flipped horizontally).}
\label{fig:random_hair}
\end{figure}

We define identity as a unique combination of facial geometry, texture (albedo and displacement), eye color and hair style.
For each identity, we render a number of images where all other parameters are varied to encourage the network to learn robust embeddings. 
While hair style \textit{can} change for an individual, most people maintain similar hair style (for both facial and head hair) which makes hair style an important cue for the person's identity. 
Consequently, for the same identity, we randomize only the color, density and thickness of the hair (see \autoref{fig:random_hair} for examples), and the hair style is only changed when the added head-wear is not compatible with the original hair style to avoid intersection (e.g., third image of top row in \autoref{fig:id_and_acc}).
For sampling facial geometry, texture and eye color we follow \cite{2021_FakeItMakeIt}.

For a given identity, we sample different accessories including clothing, make-up, glasses, face-wear (e.g., face masks) and head-wear (e.g., hats).
After selecting the clothing randomly from the digital wardrobe, other accessories are added with probability $p=\{0.15, 0.15, 0.01, 0.15\}$ respectively.
We also add hands and secondary faces with a small probability ($p=0.01$) to simulate the case when (1) the face is occluded by hands and when (2) there are multiple faces in the image.
\autoref{fig:id_and_acc} shows examples of the sampled identities rendered with different sets of accessories.

\begin{figure}[t]
\begin{center}
\includegraphics[width=\linewidth]{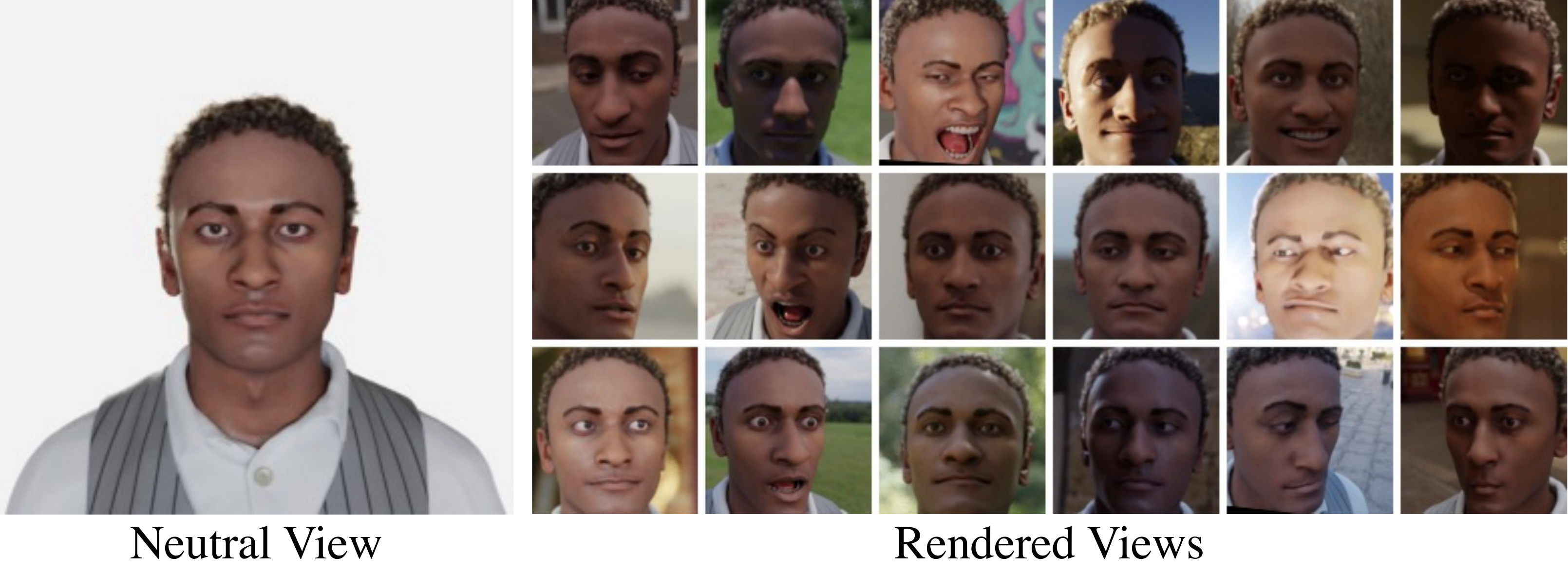}
\end{center}
\caption{Examples of images rendered for the same identity and accessory setup. The same face can look very different depending on the pose, expression, environment (lighting and background) and camera, encouraging the network to learn robust embedding.}
\label{fig:examples_of_18}
\end{figure}

For each accessory setup, we vary the pose, expression, camera and environment (lighting and background) to render multiple images.
The camera is rotated around the face, both horizontally and vertically. 
Horizontal angle is sampled from a truncated zero-mean normal distribution with support $\theta_\text{hori} \in [-90^\circ, 90^\circ]$. 
The variance is set such that the probability density $p(\theta_\text{hori}=90^\circ)$ equals to $10^{-3} \times p(\theta_\text{hori}=0^\circ)$. 
Vertical angle is sampled from a similar truncated normal distribution with support $\theta_\text{vert} \in [-30^\circ, 30^\circ]$ and $p(\theta_\text{vert}=30^\circ)=10^{-3} \times p(\theta_\text{vert}=0^\circ)$. 
This allows us to render a wide range of poses while making sure that frontal views are rendered more often. 
Lastly, the face is randomly translated within the viewing frustum to add additional perspective distortion. 
For pose, expression, and environment sampling, we follow \cite{2021_FakeItMakeIt}. 
\autoref{fig:examples_of_18} shows the impact of varying the pose, expression, environment and camera for the same identity and accessory setup.

\begin{figure}[t]
\begin{center}
\includegraphics[width=1.0\linewidth]{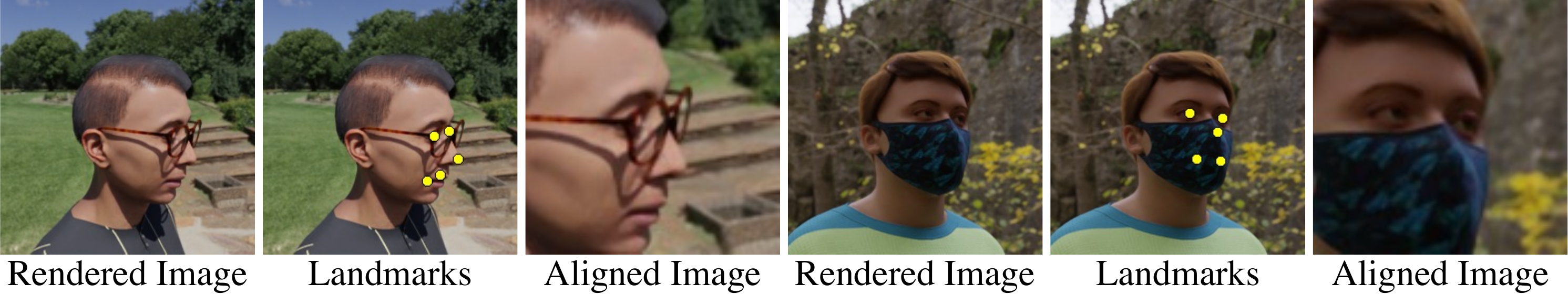}
\end{center}
\caption{For synthetic faces, it is trivial to extract the locations of ground-truth facial landmarks (e.g., eyes, nose-tip and mouth corners) and align the crop around the face. This enables robust face alignment, even when some of the landmarks are not visible.
}
\label{fig:face_alignment}
\end{figure}

\noindent
\textbf{Face alignment.} 
The input to the face embedding network should be an aligned crop around the face. 
Instead of detecting facial landmarks using pre-trained DNNs (such as MTCNN~\cite{2016_mtcnn} and RetinaFace~\cite{2020_retinaface}), we align the faces using the \textit{ground truth} landmarks (see \autoref{fig:face_alignment}), which enable robust alignment even when some landmarks are not visible.

\noindent
\textbf{Limitations.}
The face generation pipeline~\cite{2021_FakeItMakeIt} we build upon has a number of limitations resulting in domain-gap to real face images. Particularly relevant to face recognition is that we cannot generate the same person at different ages. 
While we simulate aging to some extent by randomizing the color, density and thickness of the hair (as hair typically becomes grayer, sparser and thinner during aging), more work should be done to faithfully simulate aging.
Lack of coverage (e.g., no jewelry and tattoos) may also mean that the distribution of the synthetic data does not match reality.

\subsection{Dataset Statistics}
\label{sec:method3}

The proposed dataset consists of two parts. 
The \textbf{first part} contains 720K images with 10K identities. 
For each identity, 4 different sets of accessories are sampled and 18 images are rendered for each set (i.e., 72 images-per-identity). 
Since many views of the same face are available, the network can learn embedding that is robust to the changes in accessories, camera, pose, expression, and environment. 
The \textbf{second part} contains 500K images with 100K identities. 
For each identity, only one set of accessories is sampled and only 5 images are rendered. 
This part was added to substantially increase the total number of identities with small rendering cost.
Ensuring sufficient number of identities is important since the network should learn to distinguish between similar-looking faces of different identities. 
We show in the experiments that mixing the two parts leads to better accuracy than using one of them (\autoref{table:mix_large_ids}).

\begin{figure}[t]
\begin{center}
\includegraphics[width=\linewidth]{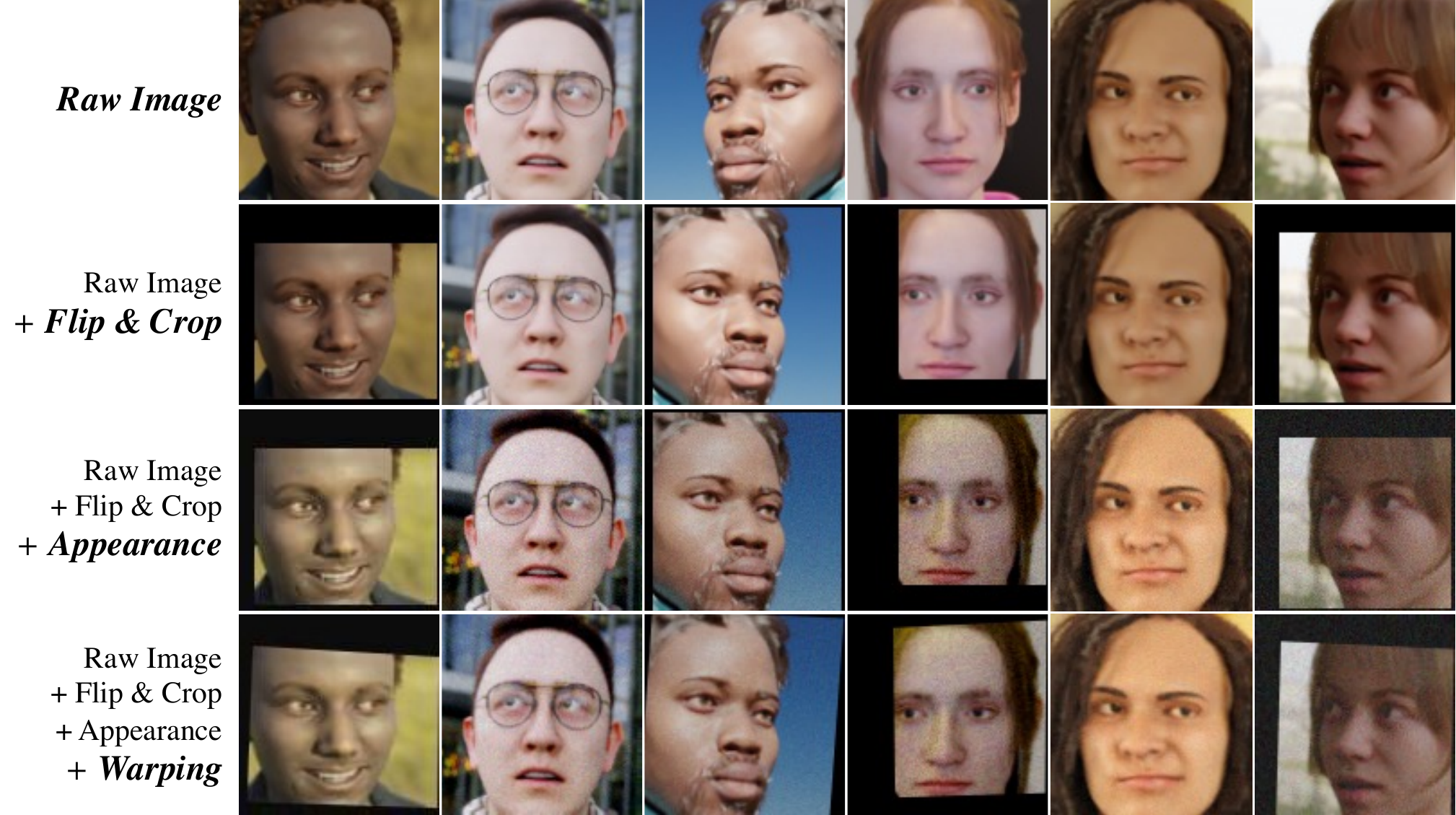}
\end{center}
\caption{Synthetic face images at different stages of data augmentation. Aggressive augmentation helps to simulate effects such as motion blur and distortion common in real-world images and thus improve the robustness of DNNs trained on synthetic images.}
\label{fig:augmentation}
\end{figure}

\subsection{Data Augmentation}
\label{sec:method4}

The quality of in-the-wild face images can vary significantly. 
Certain parts of the face may be occluded, and the images are subject to distortion and noise that are specific to each camera. 
As our synthetic faces are rendered with controlled quality using a perfect pinhole camera, aggressive data augmentation is needed to reduce the synthetic-to-real domain-gap. 
We first apply random horizontal flipping and cropping, following \cite{2022_FR_adaface}. 
Then, we apply two sets of augmentations - appearance and warping. 
\autoref{fig:augmentation} shows training images with these augmentations. Note that we apply the data augmentation on-the-fly during training, i.e., each epoch sees different random augmentations.
For each type of augmentation, we indicate its probability $p$ to be applied on a sample image. 

\noindent
\textbf{Appearance augmentation.} We apply random Gaussian blur ($p=0.05$) and Gaussian noise ($p=0.035$). By applying the Gaussian blur along a random direction using an anisotropic covariance, we also simulate motion blur ($p=0.05$). Brightness, contrast, hue and saturation are randomized with $p=\{0.15, 0.3, 0.1, 0.1\}$.
Images are converted into grayscale with $p=0.01$. Lastly, the image quality is randomized by downsampling-and-upsampling ($p=0.01$) and JPEG compression ($p=0.05$). 

\noindent
\textbf{Warping augmentation.} Warping is performed by randomly shifting the four corners of the image. Firstly, the aspect ratio is randomized with $p=0.1$. Then, all images undergo random scaling, rotation and shift. Lastly, the four corners are shifted differently for additional distortion.


\section{Experimental Setup}
\label{sec:exp_setup}

\noindent
\textbf{Implementation details.} 
Synthetic faces are rendered using Cycles renderer~\cite{cycles}, with 256 samples per pixel. 
The rendering of the full dataset took approximately 10 days, using 300 NVIDIA M60 GPUs. 
The images are rendered at $256 \times 256$ resolution, and the aligned crop around the face is resized into $112 \times 112$. 
We use ResNet-50~\cite{2016_resnet} backbone for the experiments in \autoref{sec:exp1}, \ref{sec:exp2} and \ref{sec:exp3}. 
For comparison against the state-of-the-art methods in \autoref{sec:exp4}, we use their encoder architecture to ensure fair comparison. 
For all experiments, the networks are implemented with PyTorch~\cite{PyTorch} and are trained for 40 epochs using SGD. 
The batch size is set to 256 and the networks are trained on four NVIDIA P100 GPUs. 
We follow the learning rate scheduling of \cite{2021_FR_synface}, and use the training loss from \cite{2022_FR_adaface}. 
Note that all networks are trained from scratch (not pre-trained on, e.g., ImageNet~\cite{2009_imagenet}), to make sure that no real images are used.

\noindent
\textbf{Evaluation protocol.} 
Following state-of-the-art methods~\cite{2020_FR_curricularface,2021_FR_dcq,2021_FR_magface,2021_FR_spherical,2022_FR_adaface}, we report the face verification accuracy on five benchmark datasets - LFW~\cite{2008_FRD_LFW}, CFP-FP~\cite{2016_FRD_CFPFP}, CPLFW~\cite{2018_FRD_CPLFW}, AgeDB~\cite{2017_FRD_AGEDB} and CALFW~\cite{2017_FRD_CALFW}. 
LFW contains 6,000 pairs of in-the-wild face images. 
CFP-FP and CPLFW have \textit{larger pose variation} (CFP-FP specifically compares frontal views to profile views). 
AgeDB and CALFW have \textit{larger age variation}.


\begin{table*}[t]
\footnotesize
\setlength{\tabcolsep}{3.0pt}
\begin{center}
\begin{tabular}{l|l|ccccc|c}
\toprule
Experiment & Method & LFW & CFP-FP & CPLFW & AgeDB & CALFW & Avg \\
\midrule
\multirow{4}{*}{Data augmentation}
& No augmentation
& 88.07 & 70.99 & 66.73 & 60.92 & 69.23 & 71.19 \\
& Augmentation from AdaFace \cite{2022_FR_adaface}
& 90.12 & 76.41 & 71.33 & 67.17 & 74.13 & 75.83 \\
& Ours (appearance)
& 94.32 & 80.00 & 74.83 & 75.82 & 76.92 & 80.38 \\
& Ours (appearance + warping)
& \textbf{94.55} & \textbf{84.86} & \textbf{77.08} & \textbf{76.97} & \textbf{77.20} & \textbf{82.13} \\
\bottomrule
\end{tabular}
\end{center}
\caption{The proposed aggressive data augmentation significantly improves the accuracy across all datasets. 
}
\label{table:data_preprocessing}
\end{table*}

\begin{table*}[t]
\footnotesize
\setlength{\tabcolsep}{3.0pt}
\begin{center}
\begin{tabular}{l|l|ccccc|c}
\toprule
Experiment & Method & LFW & CFP-FP & CPLFW & AgeDB & CALFW & Avg \\
\midrule
\multirow{2}{*}{Accessory sampling}
& Fix accessory & 93.50 & \textbf{82.16} & \textbf{75.75} & 73.05 & 73.83 & 79.66 \\
& Randomize accessory & \textbf{94.23} & 82.04 & 75.18 & \textbf{76.43} & \textbf{77.22} & \textbf{81.02} \\
\midrule
\multirow{3}{*}{Pose sampling}
& Minimize horizontal angle
& 93.42 & 67.19 & 66.48 & \textbf{76.78} & \textbf{77.22} & 76.22 \\
& Minimize vertical angle 
& 93.67 & 81.13 & 74.57 & 76.57 & 76.68 & 80.52 \\
& Random pose
& \textbf{94.23} & \textbf{82.04} & \textbf{75.18} & 76.43 & \textbf{77.22} & \textbf{81.02} \\
\midrule
\multirow{3}{*}{Texture sampling}
& 50
& 89.63 & \textbf{75.04} & 69.72 & 69.47 & 70.10 & 74.79 \\
\multirow{3}{*}{\footnotesize (\# textures to select from)}
& 100
& \textbf{90.83} & 74.84 & \textbf{70.30} & 70.62 & \textbf{70.57} & \textbf{75.43} \\
& 150
& 90.03 & 73.01 & 69.63 & \textbf{71.48} & 70.27 & 74.89 \\
& 200
& 89.82 & 73.37 & 69.37 & 71.45 & 70.50 & 74.90 \\
\bottomrule
\end{tabular}
\end{center}
\caption{Dataset composition experiments to study how the sampling of each attribute affects the accuracy.}
\label{table:data_composition}
\end{table*}

\section{Experiments}
\label{sec:exp}

We run a series of experiments to demonstrate the usefulness of the proposed dataset. 
Subsection \ref{sec:exp1} compares different data augmentations. 
In \autoref{sec:exp2}, we train the network on various different subsets of the full dataset to understand how each attribute sampling in rendering affects the accuracy. 
In \autoref{sec:exp3}, we show that our synthetic faces can be used in conjunction with a small number of real faces to substantially improve the accuracy. 
Lastly, we provide comparison against the state-of-the-art methods in \autoref{sec:exp4}.

\subsection{Data Augmentation}
\label{sec:exp1}

In \autoref{sec:method4}, we introduced \textit{appearance} and \textit{warping} augmentations. 
As shown in \autoref{table:data_preprocessing}, both lead to significant improvement across all datasets. 
We also compare against the augmentation used by AdaFace~\cite{2022_FR_adaface}, which includes horizontal flipping, cropping and \textit{mild} color augmentation.
For our synthetic face images which are free of imperfection, more aggressive data augmentation is needed to reduce the domain-gap. 
Notice that the warping augmentation improves the performance especially for the large-pose-variation datasets (CFP-FP and CPLFW). 

\subsection{Dataset Composition}
\label{sec:exp2}

Having full control over the rendering pipeline, we can create a dataset with desired statistics to study how each attribute affects the face recognition accuracy.
The results are provided in \autoref{table:data_composition}. 

\noindent
\textbf{Accessory sampling.} 
For 10K synthetic identities, we sampled 4 accessory setups and rendered 18 images for each setup (i.e., 720K images in total). 
These 18 images have variations in pose, expression, camera, and environment (see \autoref{fig:examples_of_18}). 
From this, we can create a subset of 180K images by selecting 18 images per ID with \textit{fixed accessory}. 
Similarly, we can select 18 images \textit{randomly} so that images with different accessories are used during training. 
When randomizing the accessories, we also randomized the color, thickness and density of the hair to simulate aging (\autoref{fig:random_hair}). 
As a result, the accuracy is improved especially for the large-age-variation datasets (AgeDB and CALFW). 
For CFP-FP and CPLFW, which has smaller age gap (i.e., positive pairs capture the identity at similar age), fixing the accessory and hair leads to slightly better accuracy.

\noindent
\textbf{Pose sampling.} 
Similar to the accessory sampling, we can select 18 images for each of the 10K identities by selecting the ones with the smallest horizontal/vertical angles.
Then, we can compare them against the 18 images selected randomly. 
For the randomly selected images, the standard deviation in horizontal and vertical angles were $(\sigma_\text{hori}, \sigma_\text{vert}) = (24.13^\circ, 9.20^\circ)$. 
For the images with the smallest horizontal/vertical angles, they were $(4.71^\circ, 8.06^\circ)$ and $(22.02^\circ, 1.72^\circ)$ respectively. 
As shown in Row 3-5 in \autoref{table:data_composition}, increasing the variation in horizontal and vertical angles improved the accuracy especially for the large-pose-variation datasets (CFP-FP and CPLFW). 
For AgeDB and CALFW, which consists mainly of \textit{frontal} faces, the accuracy was similar.

\begin{figure}[t]
\begin{center}
\includegraphics[width=\linewidth]{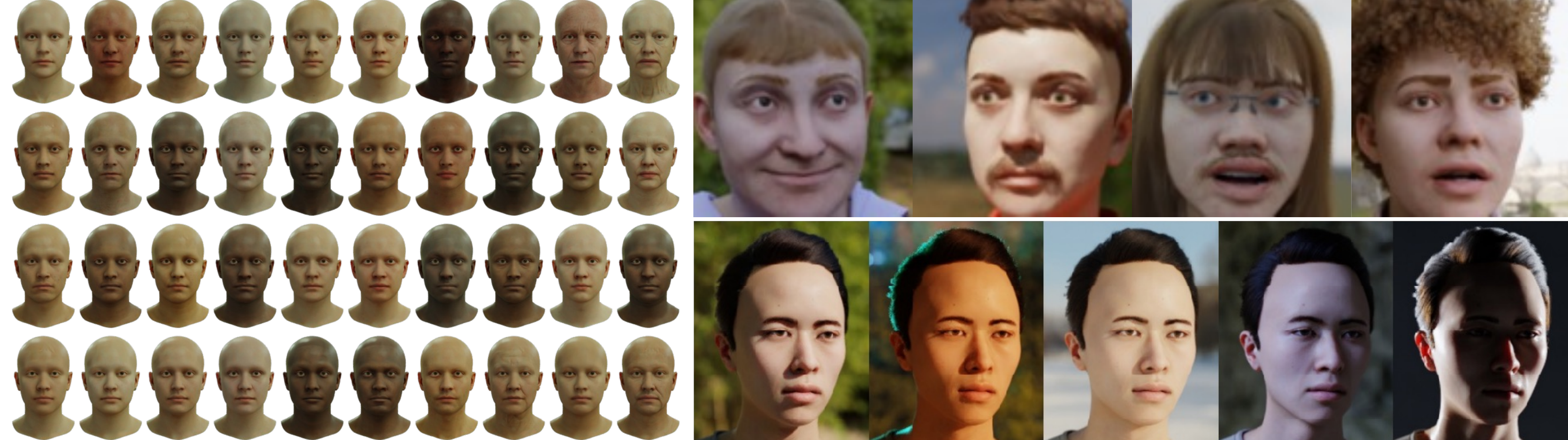}
\end{center}
\caption{Left: 40 textures selected randomly from the texture library. The library covers diverse skin color and age. Right top row: various identities (facial geometry and hair style) sampled with the same texture. Right bottom row: same identity with the same texture under different environments (taken from \cite{2021_FakeItMakeIt}). With large variations in geometry, hair style and environments, rich appearance variation could be achieved with limited textures.
}
\label{fig:num_textures}
\end{figure}

\noindent
\textbf{Texture sampling.} 
While we can create infinite number of unique facial geometries, the texture is sampled from a library built from 208 scans of real human faces (obtained with consent). 
Since we generated 110K identities in total, many of them share the same texture. 
To see how the number of textures affects the accuracy, we created a dataset of 1200 identities with $N$ textures, by generating $1200/N$ identities for each texture. 
As shown in Row 6-9 of \autoref{table:data_composition}, increasing the number of textures did not lead to a meaningful improvement in the accuracy.
This is contrary to the intuition that small number of textures and lack of texture generative model are limitations of synthetic data for face recognition. 
We believe that the appearance variability is a combination of geometry, texture, hair, accessories, environment and image quality.
In \autoref{fig:num_textures}, we show that (1) the texture library already covers diverse skin color and age, (2) an arbitrary number of unique identities can be generated with the same texture, and (3) skin appearance is greatly affected by the environment.
Also, the image quality for face recognition task is in general limited due to low resolution and data augmentation. 
Thus, the contribution of texture variation is likely less important than that of geometry and environment.

\begin{table}[t]
\footnotesize
\setlength{\tabcolsep}{2.0pt}
\begin{center}
\begin{tabular}{c|ccccc|c}
\toprule
\underline{\# IDs} $\times$ \# images/ID  & LFW & CFP-FP & CPLFW & AgeDB & CALFW & Avg \\
\midrule
\underline{10K} $\times$ 50 + \underline{0} $\times$ 5 & 94.38 & 84.07 & 76.53 & 75.93 & 76.72 & 81.53 \\
\underline{8K} $\times$ 50 + \underline{20K} $\times$ 5 & 94.80 & 84.79 & 77.52 & 76.47 & 77.65 & 82.24 \\
\underline{6K} $\times$ 50 + \underline{40K} $\times$ 5 & 95.22 & \textbf{85.24} & 77.15 & 77.52 & 78.32 & 82.69 \\
\underline{4K} $\times$ 50 + \underline{60K} $\times$ 5 & \textbf{95.45} & 84.83 & 77.70 & \textbf{77.68} & \textbf{79.10} & \textbf{82.95} \\
\underline{2K} $\times$ 50 + \underline{80K} $\times$ 5 & 94.82 & 84.09 & \textbf{77.75} & 77.55 & 78.37 & 82.51 \\
\underline{0} $\times$ 50 + \underline{100K} $\times$ 5 & 94.45 & 83.34 & 76.77 & 76.33 & 77.28 & 81.64 \\
\bottomrule
\end{tabular}
\end{center}
\caption{Number of IDs and number of images/ID should both be high to learn diverse and robust embedding. Mixing two datasets with large/small number of images/ID can be an efficient way of satisfying both. The overall accuracy becomes higher than relying on one of the two datasets.}
\label{table:mix_large_ids}
\end{table}
    
\begin{figure*}[t]
\begin{center}
\includegraphics[width=\linewidth]{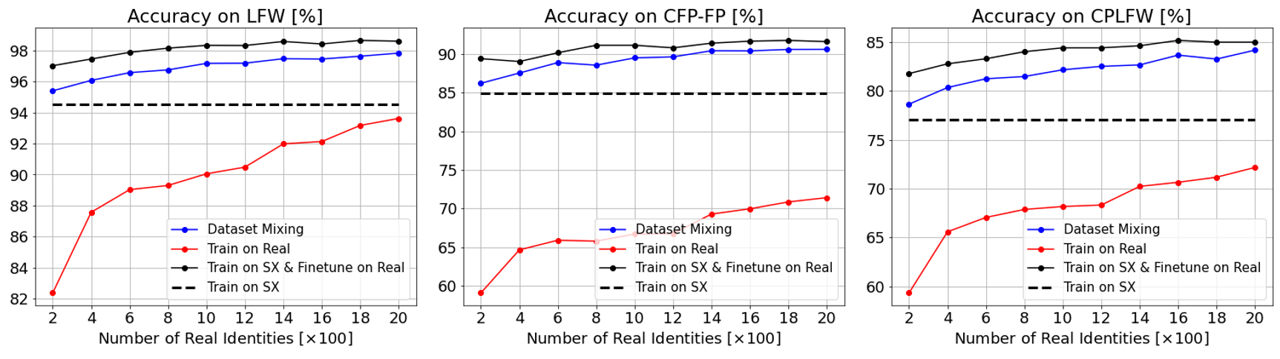}
\end{center}
\caption{Comparison between training with our synthetic data only (black dashed line), with small amount of real data only (red line), with the mixture of the two (blue line), and pre-training on synthetics and fine-tuning with the real data (black line). The number of real identities varies from 200 to 2000, and 20 images are sampled for each identity.
When only a small number of real face images are available (e.g., due to ethical issues), the proposed synthetic dataset can substantially improve the accuracy.}
\label{fig:mix_real}
\end{figure*}

\noindent
\textbf{Balance between \# IDs and \# images/ID.} 
Ensuring large number of IDs is important for learning diverse discriminative embedding. 
On the other hand, large number of images per ID (referred to as \textit{images/ID}) is needed for learning robust embedding (that is not affected by the changes in pose, accessories, expressions, camera and environment).
Mixing the two datasets with different number of images/ID can be considered as an efficient way of getting the best of both. 
This also simulates the long-tailed distribution of real face datasets (i.e., most identities have small number of images). 
The result in \autoref{table:mix_large_ids} shows that mixing the two datasets leads to better accuracy than using one of them.

\subsection{Mixing with Real Faces}
\label{sec:exp3}

The main problems associated with large-scale real face datasets are \textit{ethical issues}, \textit{label noise} and \textit{data bias}. 
In this study, we assume a scenario where a small number of real face images are collected with consent. 
For small number of images, it would also be possible to remove (or reduce) the label noise and data bias.

For synthetic data, we used 10K identities with 72 images per identity. For real face images, we varied the number of identities from 200 to 2000, with 20 images sampled for each identity (the identities and images were sampled randomly from CASIA-WebFace~\cite{2014_FRD_casiawebface}).

We first tried training only on the synthetic data. Secondly, we tried training only on the real data. Then, we explored two different strategies for using both real and synthetic images: (1) dataset mixing and (2) pre-training on synthetic data and fine-tuning on the real data. For fine-tuning, we reduced the learning rate by 1/10 for the prediction head, and 1/100 for the encoder to avoid catastrophic forgetting. The results are provided in \autoref{fig:mix_real}.

When the network is trained only on a small number of real face images, the accuracy is worse than the network trained only on our synthetic dataset.
Both dataset mixing and pre-training can lead to significantly higher accuracy, especially for the large-pose-variation datasets (CFP-FP and CPLFW). 
Compared to dataset mixing, pre-training on synthetics followed by fine-tuning on real images led to better accuracy.
This can be due to the imbalance between the number of images (we use 720K synthetic images, and a lot fewer real images).

\subsection{Comparison to the State-of-the-Art}
\label{sec:exp4}

\begin{table*}[]
\footnotesize
\setlength{\tabcolsep}{2.9pt}
\begin{center}
\begin{tabular}{c|c|c|ccccc|cc}
\toprule
\multirow{2}{*}{Method} & \# Synthetic images & \# Real images
& \multirow{2}{*}{LFW} & \multirow{2}{*}{CFP-FP} & \multirow{2}{*}{CPLFW} & \multirow{2}{*}{AgeDB} & \multirow{2}{*}{CALFW} & \multirow{2}{*}{Avg} & \multirow{2}{*}{Avg$^\dagger$}\\
& {\footnotesize (\# IDs $\times$ \# imgs/ID)} & {\footnotesize (\# IDs $\times$ \# imgs/ID)}
& & & & & & \\
\midrule
SynFace \cite{2021_FR_synface} & 500K {\small (10K$\times$50)} & 0 &
91.93 & 75.03 & 70.43 & 61.63 & 74.73 & 74.75 & 79.13\\
Ours & 500K {\small (10K$\times$50)} & 0 
& 95.40 & 87.40 & 78.87 & 76.97 & 78.62 & 83.45 & 87.22\\
Ours & 1.22M {\small (10K$\times$72+100K$\times$5)} & 0 
& \textbf{95.82} & \textbf{88.77} & \textbf{81.62} & \textbf{79.72} & \textbf{80.70} & \textbf{85.32} & \textbf{88.74}\\
\midrule
SynFace \cite{2021_FR_synface} & 500K {\small (10K$\times$50)}  & 40K {\small (2K$\times$20)} &
97.23 & 87.68 & 80.32 & 81.42 & 85.08 & 86.35 & 88.41 \\
Ours & 500K {\small (10K$\times$50)}  & 40K {\small (2K$\times$20)} & 
99.05 & 94.01 & 87.27 & 89.77 & 90.08 & 92.04 & 93.44 \\
Ours & 1.22M {\small (10K$\times$72+100K$\times$5)} & 40K {\small (2K$\times$20)} &
\textbf{99.17} & \textbf{94.63} & \textbf{88.10} & \textbf{90.50} & \textbf{90.97} & \textbf{92.67} & \textbf{93.97}\\
\bottomrule
\end{tabular}
\end{center}
\caption{Comparison to SynFace using the same encoder architecture (LResNet50E-IR~\cite{2021_FR_synface}). For both scenarios - training only on synthetic faces \& using a small number of real faces - we significantly outperform SynFace across all datasets. Avg$^\dagger$ shows average of LFW, CFP-FP and CPLFW, excluding the large-age-variation datasets.}
\label{table:synface_comparison}
\end{table*}


\begin{table*}[t]
\footnotesize
\setlength{\tabcolsep}{2.9pt}
\begin{center}
\begin{tabular}{c|c|c|ccccc|cc}
\toprule
Method & \# Synthetic images & \# Real images
& LFW & CFP-FP & CPLFW & AgeDB & CALFW & Avg & Avg$^\dagger$ \\
\midrule
Ours (SX best) & 1.22M & 0 
& 96.17 & 89.81 & 82.23 & 81.10 & 82.55 & 86.37 & 89.40
\\
Ours (SX+Real best) & 1.22M & 120K
& 99.33 & 95.93 & 89.47 & 91.55 & 91.78 & 93.61 & 94.91
\\
\hline
SV-AM-Softmax \cite{2018_FR_SV-AM-Softmax} & \multirow{6}{*}{0} & \multirow{6}{*}{5.8M}
& 99.50 & 95.10 & 89.48 & 95.68 & 94.38 & 94.83 & 94.69
\\
SphereFace \cite{2017_FR_SphereFace} &   & 
& 99.67 & 96.84 & 91.27 & 97.05 & 95.58 & 96.08 & 95.93
\\
CosFace \cite{2018_FR_CosFace} & & 
& 99.78 & 98.26 & 92.18 & \textbf{98.17} & \textbf{96.18} & 96.91 & 96.74 \\
ArcFace \cite{2019_FR_ArcFace} & &
& 99.81 & 98.40 & 92.72 & 98.05 & 95.96 & 96.99 & 96.98 \\
MagFace \cite{2021_FR_magface} & &
& \textbf{99.83} & 98.46 & 92.87 & \textbf{98.17} & 96.15 & 97.10 & 97.05 \\
AdaFace \cite{2022_FR_adaface} & &
& 99.82 & \textbf{98.49} & \textbf{93.53} & 98.05 & 96.08 & \textbf{97.19} & \textbf{97.28}
\\
\bottomrule
\end{tabular}
\end{center}
\caption{Comparison to the state-of-the-art methods trained on real face images (MS1MV2~\cite{2019_FR_ArcFace}). We use the same backbone (ResNet100) for fair comparison. By only using 120K real face images (2\% of MS1MV2~\cite{2019_FR_ArcFace}), we achieve accuracy that is comparable to the methods trained on millions of real face images. Since we do not model aging explicitly, our accuracy is worse for large-age-variation datasets (AgeDB and CALFW). Avg$^\dagger$ shows average of LFW, CFP-FP and CPLFW, and on these, we outperform \cite{2018_FR_SV-AM-Softmax} and are similar to \cite{2017_FR_SphereFace}.
}
\label{table:sota_comparison}
\end{table*}

\noindent
\textbf{Comparison to SynFace.} 
SynFace~\cite{2021_FR_synface} is the current state-of-the-art for \textit{face recognition model trained on synthetic faces}. 
They used DiscoFaceGAN~\cite{2020_discofacegan} to generate 500K synthetic faces (10K identities \& 50 images/ID). 
To ensure a fair comparison, we trained the same encoder (LResNet50E-IR) with same number of images.
We also trained using our full dataset (1.22M images). 
The results are provided in Row 1-3 of \autoref{table:synface_comparison}. 
In the second scenario, we additionally used 40K real face images from CASIA-WebFace~\cite{2014_FRD_casiawebface}. 
While SynFace \textit{mixed} their synthetic dataset with the real faces, we instead adopted the two-stage method of pre-training and fine-tuning as discussed in \autoref{sec:exp3}.
The results are provided in Row 4-6 of \autoref{table:synface_comparison}.

For both scenarios, we significantly outperform SynFace across all datasets.
This suggests that our \textit{rendered} synthetic faces are better than \textit{GAN-generated} faces for learning face recognition. 
While GANs like \cite{2020_discofacegan} can generate realistic face images, the data they generate is not ideal for face recognition, due to following reasons: \textbf{(1) Identity change.}
While \cite{2020_discofacegan} is \textit{encouraged} to preserve the identity when changing other latent variables, there is no guarantee that the identity will be preserved during data generation.
\textbf{(2) Geometric inconsistency.}
As pointed out by \cite{2022_gram}, the images generated by \cite{2020_discofacegan} for same identity and different poses lack 3D consistency. 
\textbf{(3) Lack of accessory change.} 
\cite{2020_discofacegan} cannot randomize accessories. 
\textbf{(4) Unresolved ethical concerns.}
Training the GAN model itself requires large-scale real face dataset.
For example, 70K images are used to train \cite{2020_discofacegan}. 
To learn to preserve identity, they also used a perceptual loss based on \cite{2017_discofacegan_perceptual}, which is trained on 3M real face images. 

In Row 2 and 3 of \autoref{table:synface_comparison}, we increase our synthetics dataset size from 500K to 1.22M, and achieve better accuracy. 
This indicates that the accuracy may not have converged yet and could be improved further by generating more synthetic data.

\noindent
\textbf{Comparison to methods trained on real faces.} 
Lastly, we compare the accuracy against the methods that are trained on real face images.
In \autoref{table:sota_comparison}, we provide the accuracy of six methods that use ResNet100 as the embedding network and MS1MV2~\cite{2019_FR_ArcFace} as the training data. 
We trained the same architecture on our synthetic dataset (Row 1). 
We also tried fine-tuning the network on a small number of real face images (Row 2). 
When trained only with the proposed synthetic dataset, the network can achieve 96.17\% on LFW. For LFW, CFP-FP and CPLFW (excluding the high-age-variation datasets), the average accuracy is 89.40\%. 
By fine-tuning the network on just 120K images ($2.0\%$ of MS1MV2), the accuracy becomes comparable to the methods trained on MS1MV2 (e.g., average accuracy on LFW, CFP-FP and CPLFW becomes higher than that of SV-AM-Softmax~\cite{2018_FR_SV-AM-Softmax}).

The performance of our method on AgeDB~\cite{2017_FRD_AGEDB} and CALFW~\cite{2017_FRD_CALFW} has a significantly larger gap than for the other datasets evaluated.
This is expected given the lack of aging simulation in our synthetic data.
We suspect that other causes of domain-gap, as described at the end of \autoref{sec:method1}, are the primary reason for the remaining performance gap for other evaluation datasets.
Reducing this domain-gap remains an area of ongoing work for our synthetic data and is likely to result in improved performance for all downstream tasks, including face recognition. We leave this as future work.

\section{Conclusion}

In this paper, we introduced a new large-scale synthetic dataset for face recognition by rendering digital faces using a graphics pipeline. 
We ran extensive experiments to study how data augmentation and various other attributes affect the accuracy. 
We demonstrated that our synthetic faces are significantly better than the GAN-generated faces for learning face recognition. 
With a small number of real face images, we achieve accuracy that is comparable to the methods trained on millions of web-crawled face images. 
We hope this dataset would be a meaningful step towards developing socially responsible face recognition models that do not depend on privacy-sensitive data obtained without consent.

{\small
\bibliographystyle{ieee_fullname}
\bibliography{egbib}
}

\end{document}